# Construction contract risk identification based on knowledge-augmented language model

Saika Wong[a], Chunmo Zheng[a], Xing Su[a, *], Yinqiu Tang[b]

[a]College of Civil Engineering and Architecture, Zhejiang University, Hangzhou, China
[b]PowerChina Huadong Engineering Corporation Limited, Hangzhou, China

Abstract

Contract review is an essential step in construction projects to prevent potential losses. However, the current methods for reviewing construction contracts lack effectiveness and reliability, leading to time-consuming and error-prone processes. While large language models (LLMs) have shown promise in revolutionizing natural language processing (NLP) tasks, they struggle with domain-specific knowledge and addressing specialized issues. This paper presents a novel approach that leverages LLMs with construction contract knowledge to emulate the process of contract review by human experts. Our tuning-free approach incorporates construction contract domain knowledge to enhance language models for identifying construction contract risks. The use of a natural language when building the domain knowledge base facilitates practical implementation. We evaluated our method on real construction contracts and achieved solid performance. Additionally, we investigated how large language models employ logical thinking during the task and provide insights and recommendations for future research.

Highlights

- The study uncovers the mechanism of “logical thinking” of large language models during construction contract risk identification.
- The construction contract knowledge-augmented framework formalizes the integration of vectorized domain knowledge into large language models.
- The construction contract risk knowledge base established in natural language facilitates its practical implementation.

* Corresponding author, E-mail address: xsu@zju.edu.cn

## 1. Introduction

Construction contracts are the foundation for relationships among project stakeholders, protecting their rights and interests throughout the project's lifespan. Contractual risks remain a long-standing and significant concern for all parties involved in a construction project, and failure to identify these risks in the contract clauses may result in disputes, posing a risk of project loss. According to a recent report (Victoria et al., 2022), the global average values and durations of disputes are $52.6M and 15.4 months, respectively.

Contract review involves several important tasks, such as identifying and modifying ambiguous clauses (Artan Ilter and Bakioglu, 2018), clarifying vaguely stated requirements (Hassan and Le, 2020; ul Hassan et al., 2020; ul Hassan and Le, 2021), and correcting inaccurately referenced specifications (Hamie and Abdul-Malak, 2018). Currently, the construction industry mainly relies on manual review due to the lack of sophisticated and reliable automated methods for identifying construction contract risks (CCRI). However, studies (Lee et al., 2019; Moon et al., 2021) have shown that this labor-intensive approach is error-prone and time-consuming. It heavily relies on the expertise and experience of contract experts who have to review numerous contract clauses within a limited timeframe.

Many researchers have suggested various methods that utilizes computer to aid the review process, which primarily involves rule-based natural language processing (NLP) methods (Kim et al., 2020) and machine learning (ML) algorithms (Choi and Lee, 2022) (Candaş and Tokdemir, 2022). However, due to the complex language comprehension required for reviewing contract clauses, the existing approaches each suffer from different limitations. For instance, rule-based NLP requires comprehensive rules to pre-defined, which is not only inflexible but also poses a significant challenge to cover all situations (Hassan and Le, 2021). In the case of applying ML algorithms (e.g., support vector machine), it requires a substantial amount of high-quality data to train a model for a specific task. This demands considerable time investment for training, as well as data collection, cleaning, and labeling. Meanwhile, its inability to effectively capture semantic and contextual information significantly limits the performance of text understanding and processing (Hassan and Le, 2021; Yang et al., 2022).

To mitigate these problems, researchers have proposed the application of language models (LMs), to enhance the capture of semantic information from contract text (Mohamed Hassan et al., 2022; Shuai, 2023). For example, Fu's research team (Fu et al., 2023) fine-tuned a language model, DeBERTa, for construction contracts semantic annotation. Meanwhile, due to the emergence of LLMs and their distinct performance, e.g., in-context learning, improved text comprehension, and the ability to summarize long paragraphs (Zhao et al., 2023), researchers have begun to consider the use of LLMs for processing contract text. Xue et al. (Xue et al., 2022) used BART to summarize construction contracts, maintaining critical information for better interpretation of the contracts.

However, LLMs pre-trained in the general domain are not directly applicable for domain-specific tasks. As highlighted in several studies, using LLMs in specialized fields often leads to poor performance due to diluted domain-specific training data and issues that arise during the training processes (Ji et al., 2023; Kim et al., 2022; Mialon et al., 2023; Peng et al., 2023). Imperfect representation learning is one such issue, where the encoders learn incorrect correlations between different parts of the training data, leading to erroneous generation that deviates from the input. Similarly, direct application of LLMs to process construction contracts may lead to unprofessional and incorrect outputs, i.e., hallucination (Huang et al., 2023). As a result, the utilization of LLMs in the construction contract management field is not as widespread as in many other fields (Cui et al., 2023).

Motivated by the aforementioned challenges and aiming to explore how to enhance LLMs to more closely emulate human experts for CCRI, we have adopted the idea of augmenting LLMs with external knowledge (Gao et al., 2022; Guu et al., 2020; Peng et al., 2023). We refer to this external knowledge, usually acquired from domain experts, as the factual knowledge and the expert knowledge. In this paper, we introduce a novel knowledge-augmented CCRI methodology. It augments an LLM with factual knowledge and the knowledge of contract experts, without fine-tuning the language model. This significantly alleviates the hallucination phenomenon in domain knowledge-intensive task. This research provides insights for utilizing LLMs with a thinking paradigm similar to that of human experts in identifying contract risks. Simultaneously, this contributes to the body of knowledge in the automated contract review, which could substantially decrease the time spent on contract reviews and minimize risks during construction projects. Although we introduce this method in the context of identifying construction contract risks, it is a versatile approach that can be incorporated into a broad range of construction contract text processing and management tasks.

# 2. Related works

## 2.1 Natural language processing in construction contract

NLP is an effective means for computers to understand and solve a wide range of tasks involving human language. Common applications include information retrieval, information extraction, text classification, text generation, question answering etc. The final goal of NLP is to enable human languages to be processed automatically by machines (Otter et al., 2019). In the past, many NLP methods and Machine Learning (ML) algorithms based on statistical and probabilistic calculations have been proposed and widely used to process and analyze text data. This field has become a popular area due to the large amount of textual data generated during construction projects in recent years (Ding et al., 2022; Yang et al., 2022; Zhang et al., 2023).

Construction contract review is a key subject of study in construction automation scenario. Automating general tasks such as contract quality review and identifying common patterns in contracts can improve efficiency and quality. The construction contract text processing can be divided into rule-based NLP and ML-based (Salama and El-Gohary, 2016). Clauses or relevant information extraction after pre-defined rule matching is a common pattern for rule-based NLP methodology (Al Qady and Kandil, 2010; Liu et al., 2014; Zhang and El-Gohary, 2016). Lee et al. (Lee et al., 2019) developed a rule-based automatic model to detect contract clauses that disadvantage contractors and extract critical information automatically. They first define the risk clauses, construct a domain lexicon, and design information extraction rules to achieve the identification of risk clauses and extraction of relevant information. Lee et al. (Lee et al., 2020) proposed rule-based NLP to build a model containing text pre-processing, syntactic analysis, and semantic analysis for checking the omission of contractor-friendly clauses in contracts. Despite the impressive results, the scalability of the method is hindered due to its dependence on the manual crafting of rules, which seems tedious and may not function efficiently with other standards contracts. Kim et al. (Kim et al., 2020) proposed a rule-based NLP approach applying syntactic rules to increase the accuracy of matching the Subject, Verb, Object (SVO) of each sentence and semantic rules to pre-defined risk categories for extracting potentially dangerous clauses from the contract documents. Padhy et al. (Padhy et al., 2021) suggested that using common phrases from exculpatory clauses as rules to extract exculpatory clauses in any contract, and eventually extract all clauses that contain one or more common phrases in the exculpatory clauses. The paper also stated that automatically extracting the desired clauses using ML requires a large amount of training data.

Obtaining high-quality training data from construction contracts is time-consuming and labor-intensive. This is the reason for most of the rule-based NLP studies opt for such approach. Studies that construct corresponding rules can yield good results in terms of accuracy and interpretability. However, their dependence on predefined vocabulary and rules diminishes the flexibility of their application, suffer from the issue of poor generalization, where some knowledge is difficult to describe through rules, and it is both impossible and ineffective to develop all possible rules. Further, the high performance comes at the expense of considerable human effort. With the advancement in the construction industry, construction contracts are becoming increasingly complex. Consequently, the need for construction contract risk identification cannot be satisfied solely by pure NLP applications. Instead, further research should focus on pre-trained neural network models (Hassan et al., 2021).

In respect to ML-based method, commonly used ML models include support vector machine (SVM) (Candaş and Tokdemir, 2022; Hassan and Le, 2021; Lee and Yi, 2017; Ul Hassan et al., 2020; Yang et al., 2022), naïve bayes (NB) (Candaş and Tokdemir, 2022; Hassan and Le, 2021, 2020; ul Hassan et al., 2020), k-nearest neighbors (KNN), and hidden Markov model (HMM) (Candaş and Tokdemir, 2022; ul Hassan et al., 2020; ul Hassan and Le, 2021). All of which have been applied simultaneously around the above-mentioned studies and compared in terms of performance. For instance, ul Hassan and Le (Hassan and Le, 2020) utilized ML to develop a dependable classification model with a recall of 95%. The model effectively distinguished between requirement and non-requirement text within contracts.

With the advancement of deep learning (DL), this technic has also been used in the analysis of construction contract. Recurrent neural network (RNN) and long short-term memory (LSTM) have been employed for this purpose. RNN is used to train a model for DB requirement classification, and the research has yielded that the reliability of the classification depends on the text vectorization method (ul Hassan and Le, 2021). Choi and Lee (Choi and Lee, 2022) proposed to apply a bi-LSTM algorithm to train a risk level ranking (RLR) model to automatically analyze the key risk clauses in the invitation to bid (ITB) at the bidding stage. In addition, the study also mentioned that although the accuracy of text classification models has improved, it does not necessarily imply that the model comprehends the text at the same semantic level as a human would.

Although ML methods offer flexibility and efficiency, most studies face a significant limitation of the availability of training data. Collecting high-quality data is crucial for training a competent model, but it poses a challenge, particularly with domain-specific documents such as construction contracts. This challenge is further amplified by the sensitive corporate data often contained within contract information, making it even more difficult to acquire of high-quality contract training data. Table 1 lists some representative studies using rule-based NLP, ML-based, and DL-based method in processing contract textual data.

| Method | Description | Task | Literature reviews |
| --- | --- | --- | --- |
| Rule-based NLP methodology (semantic mapping rules, pattern matching, lexicon-based method) | Utilized NLP and a shallow parser for the extraction of semantic knowledge from construction contract documents. | Extracting concept relations from contracts for automatic contract review and management | (Al Qady and Kandil, 2010) 9/22/23 1:14:00 PM |
| | A rule-based mechanism to map textual elements to ontology entities, enhancing the interpretation process of impact factors in the text. | Identification of impact factors in construction claim | (Niu and Issa, 2013) |
| | Employed semantic and syntactic rules to extract detrimental clauses for contractor. | Poisonous clauses identification for automatic clauses detection | (Lee et al., 2019) |
| | Used NLP techniques to transform complex and lengthy contract documents into a simplified form for accurate Subject, Verb, Object (SVO) tagging for risk prediction and management. | Potentially hazardous clauses extraction | (Kim et al., 2020) |
| | Employed a rule-based NLP methodology to build a proactive risk assessment model, analyzing unstructured text data through syntactic analysis, and semantic analysis to identify missing contractor-friendly clauses. | Contractor-friendly clauses identification for automatic clauses detection | (Lee et al., 2020) |
| | Developed an NLP model that analyzes the syntactic, lexical, and semantic aspects of natural language in construction contracts to identify exculpatory clauses automatically. | Exculpatory clauses identification | (Padhy et al., 2021) |
| ML-based methodology | Four classification models SVM, ANN, kNN, and NB are used for unstructured text combined with digital data of bid documents for bid risk prediction. | Uncertainty risks classification based on pre-bid clarification information | (Lee and Yi, 2017) |
| | Using linear and non-linear algorithms, feature extraction, transformation, and selection, with predictive models such as logistic regression (LR), decision tree (DT), and random forest (RF), to ascertain the probability of successful contract execution. | Identification of contract execution according to financial and non-financial parameters | (Valpeters et al., 2018) |
| | Four different machine learning algorithms were tested and compared for classifying contractual text into requirement and non-requirement text. | Contract requirements classification based on subcontract discipline | (Hassan and Le, 2020) |
| | Using NLP and various supervised machine learning approaches, including NB, SVM, Logistic Regression, kNN, DT, and FNN, to develop an automated framework for classifying text describing project requirements into distinct classes. | Contract requirements classification | (Ul Hassan et al., 2020) |
| | Utilizing NLP and supervised ML, including convolutional neural network and recurrent neural network, to automate subcontract drafting, categorizing requirements into three stages: design, construction, and O&M. | Contract requirements classification based on subcontract discipline | (Hassan and Le, 2021) |
| | Two models are proposed to automatically analyze critical risk clauses in Invitation to Bid (ITB) documents: a Semantic Analysis (SA) model, which is a rule-based approach using NLP to extract key risk clauses, and a Risk Level Ranking (RLR) model, which is a train-based approach that applies bi-LSTM to rank the risk impact of each clause. | ITB risk clause classification | (Choi and Lee, 2022) |
| | Several supervised ML classifiers were used to automate the identification of vague terms in construction contract conditions. | Ambiguous contract terms identification | (Candaş and Tokdemir, 2022) |
| | ML models were used to classify text in both general and particular conditions of contracts to measure the functions of contracts. | Contractual function classification | (Yang et al., 2022) |
| DL-based methodology (the primary model used was BERT) | Fine-tuned BERT to classify near-miss reports in construction projects. | Safety reports classification | (Fang et al., 2020) |
| | Develop a named entity recognition model based on bidirectional long short-term memory architecture for construction specification review. | Construction contract named entity recognition | (Moon et al., 2021) |
| | Detecting contractual risk information from construction specifications using BERT. | Contract clause classification | (Moon et al., 2022) |
| | Construction contract text summarization using different LLMs, like BART. | Contract text summarization | (Xue et al., 2022) |
| | Fine-tuned BERT to classify obligation sentences in the context of EPC contracts for mitigating the unilateral contractual change risk. | Contract clause classification | (Shuai, 2023) |
| | Fine-tuned DeBERTa for construction contracts semantic annotation to help measuring the complexity of contractual function. | Contract clause annotation | (Fu et al., 2023) |

**Table 1.** Studies of rule-based NLP, ML-based, and DL-based methods for construction contract text processing.

## 2.2 Knowledge augmented language model

In recent years, LMs have made significant progress, due to the availability of extensive datasets and the improvement of computer power. This progress has greatly influenced the field of NLP, with the emergence of several notable LLMs[1] such as GPT (Brown et al., 2020), LLaMa (Touvron et al., 2023). As the LMs have scaled up in size, they have demonstrated emergent abilities including in-context learning, step-by-step reasoning, and instruction following. These capabilities, which may not be as pronounced in smaller LMs, have enabled LLMs to excel in zero-shot and few-shot tasks. They have also showcased exceptional proficiency in more complex endeavors such as mathematical problem-solving, question answering (Zhao et al., 2023). Recent studies have highlighted the outstanding capabilities of LLMs on question answering and logic reasoning. The study (OpenAI, 2023) suggested that GPT-4 outperforms existing pre-train language models on all benchmarks, the highest accuracy up to 96.3%, for various question-answering and commonsense reasoning. Furthermore, LLMs have proven to be highly efficient on knowledge-intensive tasks and have even outperformed human experts in certain domains.

Regardless of the distinguished performance of the LLMs, its limitations lead to phenomena such as "making something out of nothing". This is the reason why it has not been widely used in many specialized fields that require a high degree of professionalism and logical rigor, such as law. This phenomenon summarized as hallucination. Hallucination can be categorized as intrinsic and extrinsic according to their nature (Liang et al., 2022). In current research, many methods have been proposed to alleviate this problem, including fine-tuning, prompt engineering, and data augmentation.

Fine-tuning is a straightforward and highly effective method of addressing the hallucination, as it involves training the model with the knowledge of a specialized domain and tailoring it to suit the desired application (Huang et al., 2023; Thoppilan et al., 2022; Tinn et al., 2023). For instance, both the Lawyer LLaMA (Huang et al., 2023) and the ChatLaw (Cui et al., 2023) fine-tuning the LLaMA model by injecting legal domain knowledge including relevant law articles related to legal issue. These modifications successfully overcame the models' inability to leverage the legal knowledge to resolve domain-specific issues. However, fine-tuning a language model, particularly one with a large number of parameters, is not a trivial and costly task. Hence, to leverage the robust capabilities of the LLMs, studies suggested that specialized methods of prompting are required to guide them in recalling the knowledge they encapsulate, a process referred to as prompt engineering. In addition, the LLMs have also shown the abilities to tackle reasoning problems by specific instructions like Chain-of-Thought (Wei et al., 2022), Tree-of-Thoughts (Yao et al., 2023).

The data augmentation on the other hand, also known as knowledge augmentation,

[1] Note: In existing research, there is currently no uniform definition of the LLM on the minimum parameter scale in either academia or industry. In this research, we mainly consider language models larger than 10B in size as LLM.

is a method that incorporates external knowledge to improve the LLMs understanding and generation capabilities. Recent works have explore leveraging the structure data in the form of knowledge graph (KG), providing external knowledge to the LLMs for inference and interpretability. Some researchers have proposed to incorporate KGs into LLMs during the pre-training stage, which can help LLMs learn knowledge from KGs (Liu et al., 2021; Yu et al., 2021). Others researchers proposed retrieving knowledge from the KGs for LLMs during the reasoning stage (Sen et al., 2023; Hu et al., 2022). Given the KGs are challenging to construct and continuously evolving, they pose difficulties for existing methods aimed at generating new facts and representing knowledge. Consequently, there are studies representing knowledge using vector embeddings that retain semantically meaningful information (Minaee et al., 2021).

Vector embeddings represent data points in a multi-dimensional space and are commonly used in ML to characterize features of objects, such as text, images, or even entities. It typically consists of a set of numbers that form a multidimensional numerical space, wherein each dimension of a vector represents a different feature or attribute. For example, in textual data, one dimension might represent the frequency of occurrence of each word. By defining different features, it is possible to represent textual data as multi-dimensional vectors. This representation has the advantage that similar content will be closer in the vector space, so that similarity calculations, such as cosine similarity, can be used to quantify the similarity of the vectors. Compared with KGs in terms of semantic search, the vector similarity search, which locates vectors closest to a given query vector, offers more accurate and contextualized searching over data. Studies have started to use vector embeddings to represent and retrieve external knowledge in an effort to improve the performance of language model (Lazaridou et al., 2022; Nye et al., 2021). Peng et al. (Peng et al., 2023) proposed an LLM-augmenter, which utilizes similarity search to retrieve external knowledge from Wikipedia, and links the retrieved knowledge to the relevant context. They reported that the augmenter significantly improves the factuality of the answers. He et al. (He et al., 2022) used Wikipedia as an external knowledge base, applying cosine similarity to select the passages most similar to the input query sentences as knowledge relevant to the question. The results of the study show that incorporating additional knowledge surpasses methods that do not. This superiority is evident in terms of common-sense reasoning and temporal reasoning, and the faithfulness of the responses increases dramatically, especially as the scale of the model increases. Knowledge is critical for equipping statistics-based models, and vector embeddings can facilitate the development of an efficient retrieval enhancement system. Capitalizing on efficient semantic search capability, this study proposes a novel method for identify risks in contract clauses by incorporating knowledge highly relevant to the risk clauses into LLMs, not only aligning the models' behaviors with the preferences of human experts enables but also increasing trustworthiness of the answers provided by the LLMs. As a result, enhances its capabilities to identify risks in contract clauses.

# 3. Methodology

In this section, we introduce the knowledge augmented language model based CCRI methodology. It consists of four parts: building a domain-specific knowledge base, retrieving project clauses, retrieving clause-review pairs, and developing appropriate prompts. Given a construction contract clause, the goal of CCRI is to identify contradictions or omissions in the context based on a given checkpoint. The overall architecture of our approach is depicted in Fig. 1.

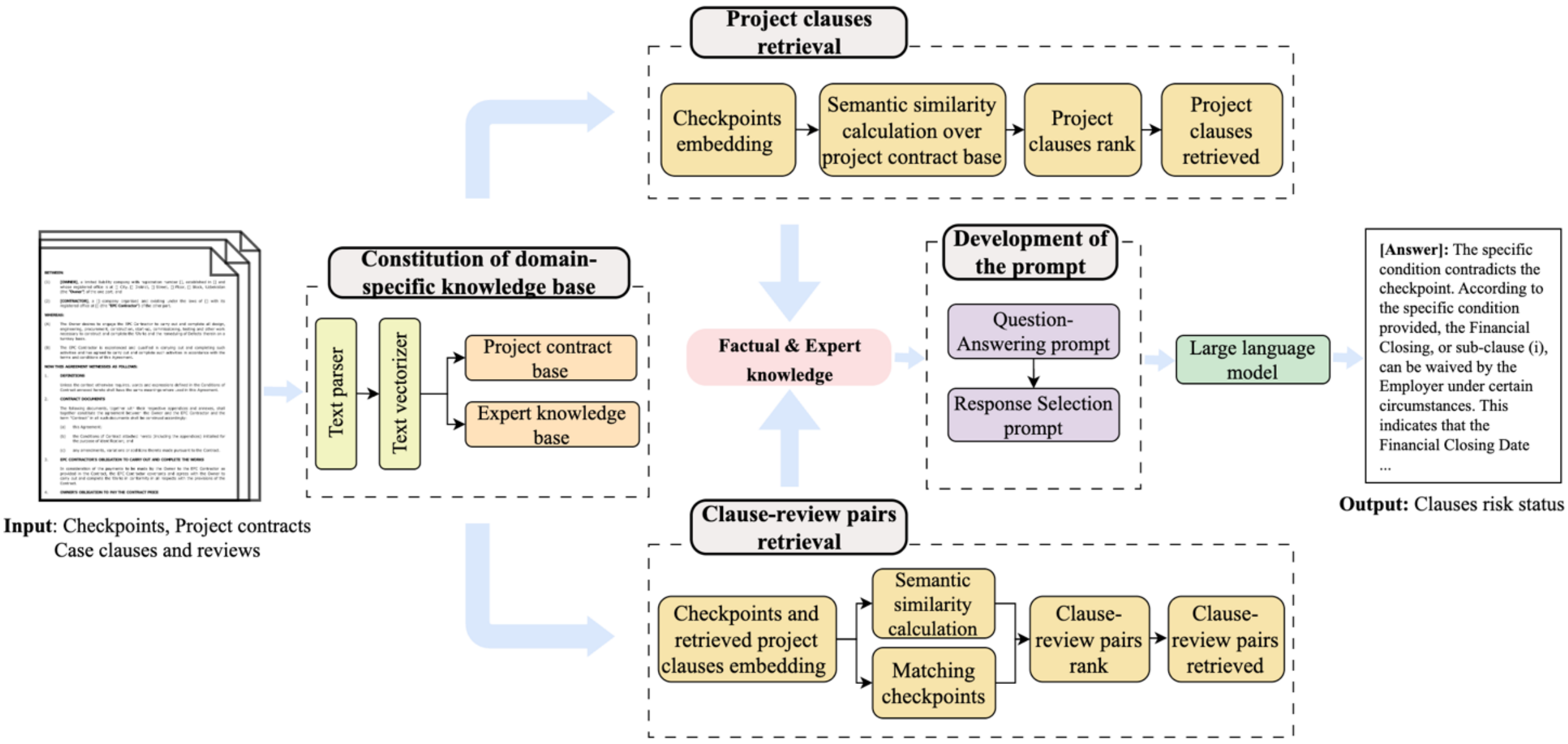

**Fig. 1.** The framework of knowledge augmented LLM for CCRI.

## 3.1 Constitution of domain-specific knowledge base

The storage of data is one of the key tasks. Before storing the contract text into the knowledge base, it needs to be vectorized. To ensure the effectiveness of semantic search, the text should be parsed into text chunks with similar attributes. In the context of the construction contracts, the text is parsed into chunks based on the content of the clauses, specifically, each section. For instance, the content of "condition precedent" and "payment" will be treated as two separate text chunks. Longer sections can be divided into shorter paragraphs while maintaining semantic coherence. Each chunk is then vectorized using a text vectorizer, represented as an n-dimensional vector of floating-point numbers. The vectorized text chunks and their corresponding embeddings are subsequently stored in the knowledge base.

The organization of data for the two knowledge bases is illustrated in Fig. 2. Each knowledge base is designed for specific purposes and has unique schemas. Both knowledge bases contain an "ID" and "Embeddings". The "ID", an integer number, is the primary key for each row of data, while the "Embeddings", a floating-point vector denotes the vector representation of each clause. The project clause base is responsible for storing and providing contract clauses for risk identification. It adds the fields "Clause_type" and "Clauses" to its schema. "Clause_type" records the type of clause, specifically the name of the section (when the subsection is omitted) or the subsection of each clause. "Clauses" represents the detailed content of each clause. For example, in subsection 4.1 "Condition Precedent", "4.1 Condition Precedent" is the "Clause_type" and the corresponding clauses "The Contract shall come into full force and effect on the Date… waived by the Owner" are the "Clauses". The "Embeddings" are then the vector representations of the relevant clauses. On the other hand, the expert knowledge base is responsible for storing the risky clauses identified in past contract reviews, along with their review results. Its schema includes three fields: "Checkpoints", "Clauses", and "Reviews". "Checkpoints" stores the contract risk identification requirements of an organization, tailored to different contract characteristics and risk preferences. Similar to the project contract base, "Clauses" stores specific clauses. However, these clauses are those that have been identified as risky in the past projects. "Reviews" records the results of the identification of such clauses, providing explanations as to why certain clauses are considered risky.

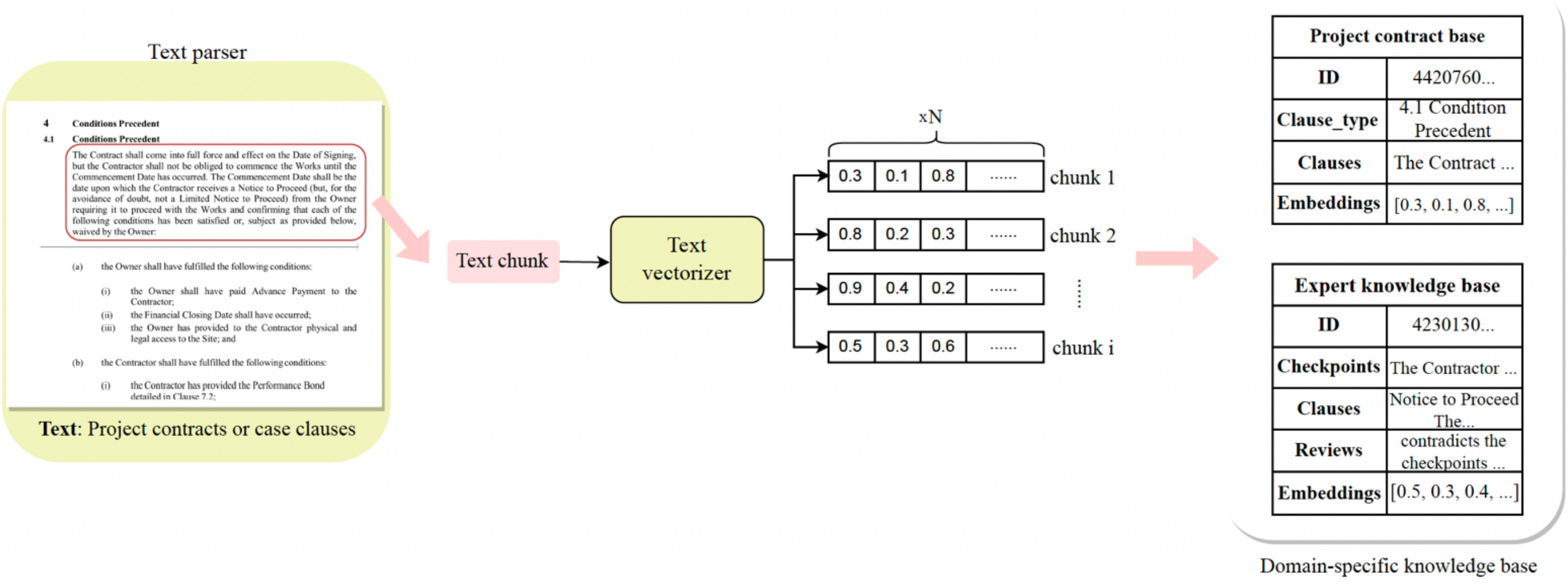

**Fig. 2.** Process of building a construction contract knowledge base.

It is essential to define the scope covered by these checkpoints, specifically delineating the key risk categories within the construction contract. In general, checkpoints should cover areas such as: Contractual Basics, Permissions and Transfers, Payment and Financials, Time and Schedule, Warranty and Performance, Liabilities and Penalties, Suspensions and Terminations, Site Condition, Claims and Variations, and Design and Subcontracting. Subsequently, based on the nature of the identification and task requirements, the composition of the checkpoints can be further specified. The extent of a checkpoint is determined by the nature of the check, which can be abstract or concrete, such as identifying specific amounts or dates. The final checkpoint is drawn based on the task requirements, which can be categorized into key point summaries, element extraction, and logical reasoning. For instance, an abstract logical reasoning checkpoint might state, "The conditions precedent should be waived by mutual agreement between the Project Company and the Contractor." Conversely, a concrete element extraction checkpoint might query, "What is the proportion of the advance payment?" The format of a checkpoint can differ, from declarative statements, as illustrated in the prior examples, to interrogative sentences, and within these sentences, it is vital to incorporate technical terms specific to the domain.

In the expert knowledge base, there is a one-to-many relationship between "Checkpoints" and "Reviews", where a single checkpoint can correspond to multiple reviews. This is because, while the descriptions of clauses in different contracts may vary, these clauses may present similar risks. For clauses with the same risks, the same checkpoint is used to review these clauses. For example, if two risk clauses "8.1 Commencement of Works; The Financial Closing has occurred in terms of the implementation agreement…" and "5.1 Notice to Proceed; The Commencement Date shall be the date stated in a written notice…" have been identified as risky for the same checkpoint "The Financial Closing Date shall have occurred before the Commencement Date" in different contracts, then these two risk clauses and their respective reviews will share the same checkpoint in the expert knowledge base.

With respect to the composition of reviews, it can be structured into three main parts: 1. Conclusion; 2. Content of the relevant clauses; 3. Rationale. For instance, "A particular provision appears to be in conflict with the checkpoint. As per section 8.1 titled 'commencement of works', specifically in sub-clause (a), it is mentioned that the obligation can be 'waived or deferred (with or without conditions) by the Employer', which prescribes that the Financial Closing should transpire prior to the Commencement Date. This implies that the requirement for the Financial Closing to precede the Commencement Date can be overruled by the Employer. Consequently, it is not definitively ascertainable that the Financial Closing Date will precede the Commencement Date." The review first directly provides a conclusion regarding the potential conflict between the clause and the checkpoint. Subsequently, it elaborates on the specific content of the relevant clause, and finally, it presents the basis for the judgement based on the clause's content.

### 3.2 Project clauses retrieval

Clause retrieval is a process designed to fetch project contract clauses that are highly associated with a given checkpoint from the pre-built project contract base (as depicted in Fig. 3). This process addresses the challenge posed by the fact that the clauses containing key information may be scattered across different locations, often involving cross-page situations. For example, Clause IV is a time provision that describes advance payment, yet the definition of advance payment is elaborated in Clause I. Clause I and Clause IV appear in different locations within the contract. The procedure begins with a manually collected checkpoint, derived from a set of checklists created by construction contract experts. These checklists focus on major inspection topics. For example, a checkpoint related to the inspection topic of "Precondition of Commencement" might include statements such as "Receipt of advance payment is not one of the commencement preconditions" or "The Owner's provision of right of access to the site is not one of the commencement preconditions".

Before searching for relevant clauses within the project contract base, the checkpoint undergoes vectorization using a text vectorizer, such as the sentence-transformers model provided by Hugging Face (Reimers and Gurevych, 2019), or the text-embedding-ada-002 model provided by OpenAI[2]. After vectorization, a semantic search is performed to retrieve the appropriate clauses. This involves calculating the vector similarity between the clauses and the checkpoint. The similarity metric can be based on either cosine similarity or Euclidean distance, both of which should yield the same ranking of results.

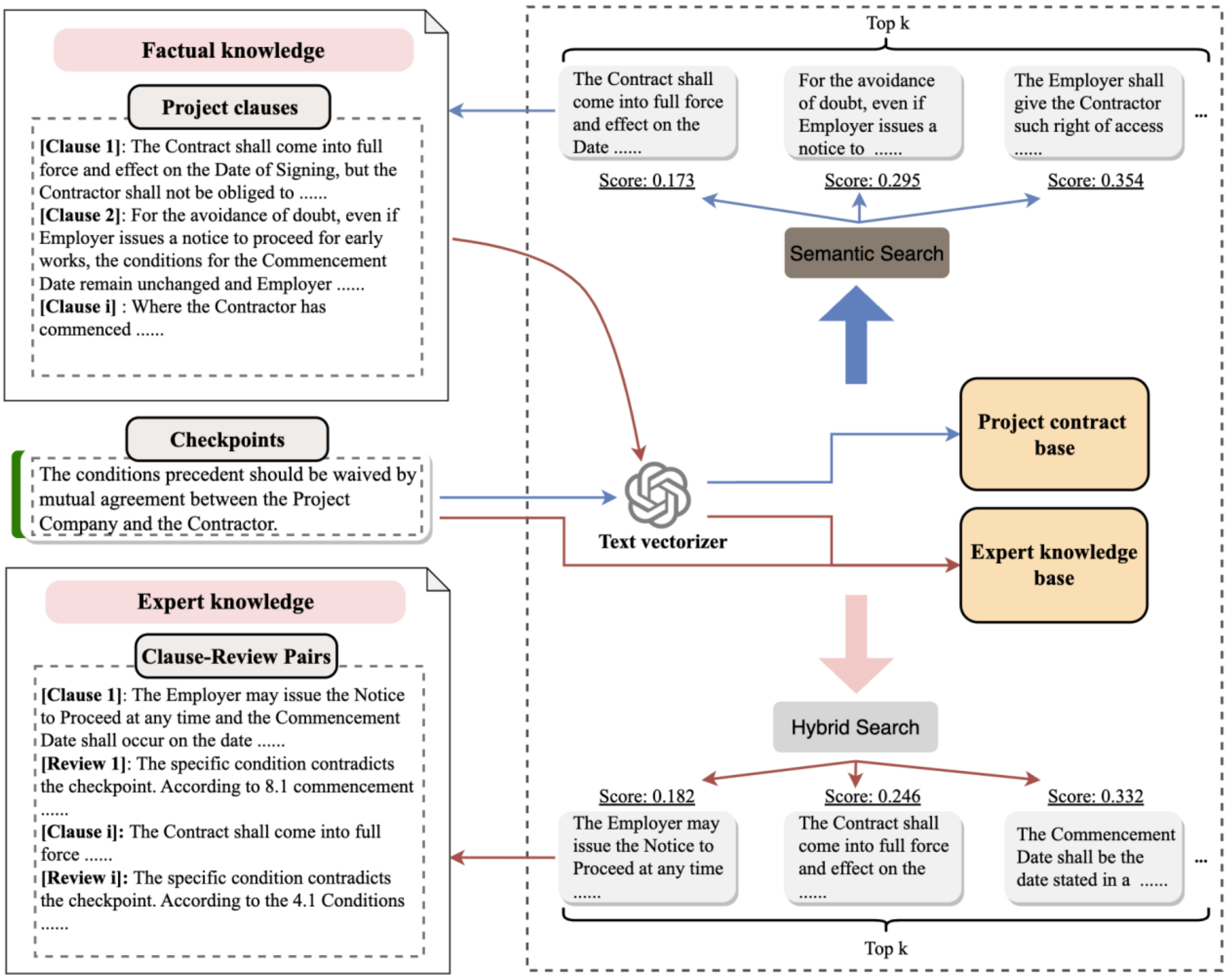

**Fig. 3.** Process of retrieving factual knowledge and expert knowledge.

### 3.3 Clause-review pairs retrieval

This step aims to acquire the case clauses and their corresponding reviews from the expert knowledge base. The case clauses are risky clauses collected from the past project contracts and the reviews are statements explaining the reason. To retrieve the clause-review pairs, we conduct a hybrid search operation, instead of pure semantic search, on the expert knowledge base using the checkpoint and the retrieved project clauses. The checkpoint retains its original text format, while the retrieved project clauses are transformed into a vector representation, composed of long segment multiple clause i. As shown in Fig. 3, after retrieving top k project clauses associated with checkpoint, according to the order of the similarity scores, we merge all the clause i into a singular text and then vectorize it.

[2] Public API available at: https://openai.com/api/.

To retrieve the case clauses relevant to the same checkpoint used to retrieve the project clauses, we first match the same checkpoint in the expert knowledge base. This narrows down the pool of clauses to those most pertinent to the checkpoint, ensuring that the retrieved case clauses possess the type of risk that the project clauses are potentially covering. After that, the top k case clauses can be retrieved by calculating the vector similarity between the project clauses and the case clauses. Finally, the most relevant case clauses and their reviews are extracted based on the similarity score.

Consider the following example: when analyzing the checkpoint "The condition precedent should be waived by mutual agreement..." and the corresponding project clauses "The Contract shall come into full force and effect on the Date...", which provide some details regarding the financial close. There are four distinct case clauses: "The Employer may issue the Notice to Proceed at any time and the Commencement Date shall occur...", "The Commencement Date shall be the date stated in a written notice to the Contractor from the Employer...", "The Company shall issue an NTP if the Conditions Precedent are defined...", and "Commencement Date refers to the date when the Contractor receives the Notice to Proceed from the Project Company...". The initial two clauses pertain to the checkpoint "The Financial Closing Date shall have occurred before the Commencement Date", whereas the latter two are associated with "The conditions precedent should only be waived by mutual agreement between the Project Company and the Contractor". Since the project clauses include references to the financial close, it is evident that using vector similarity for retrieval will lead to the extraction of clauses connected to the financial close risks, specifically the first two clauses. Nevertheless, the risk present in these two clauses arises from the financial close itself, rather than the condition precedent being waived through mutual agreement. Consequently, these clauses fail to offer insights related to mutual agreement and therefore do not contribute to the language model's understanding of the core of the checkpoint. To this end, it is necessary to first match the case clauses that have similar risks to the project clauses based on the checkpoint using vector similarity calculation. At this point, clause-review pairs related to the checkpoint can be extracted as the expert knowledge to enhance the logical reasoning of the language models.

### 3.4 Prompt development

Different prompts for the same question can produce diverse answers, which makes it challenging to find the “right” prompt to exploit the capabilities of the LLM. Leveraging the in-context learning capabilities of LLMs, we present a two-stage prompting process for generating robust, stable, and trustworthy results (as depicted in Fig. 4). The first step begins with a question-answering prompt. The content of the prompt is based on the information retrieved in the previous knowledge retrieval stages. Multiple distinct suggestions are generated by feeding the prompt into the LLM independently several times. In the second step, we use a zero-shot response selection prompt and adopt the idea of majority voting to select the most suitable and promising suggestion. The selection is based on the input context, including the checkpoint and specific conditions. This method employs self-consistent sampling rather than greedy decoding. It improves the accuracy of inference by generating a number of different answers and then applying majority voting. This response selection prompt is repeated multiple times independently, similar to the first step. The final answer for the risk identification is determined by the suggestion that receives the most votes.

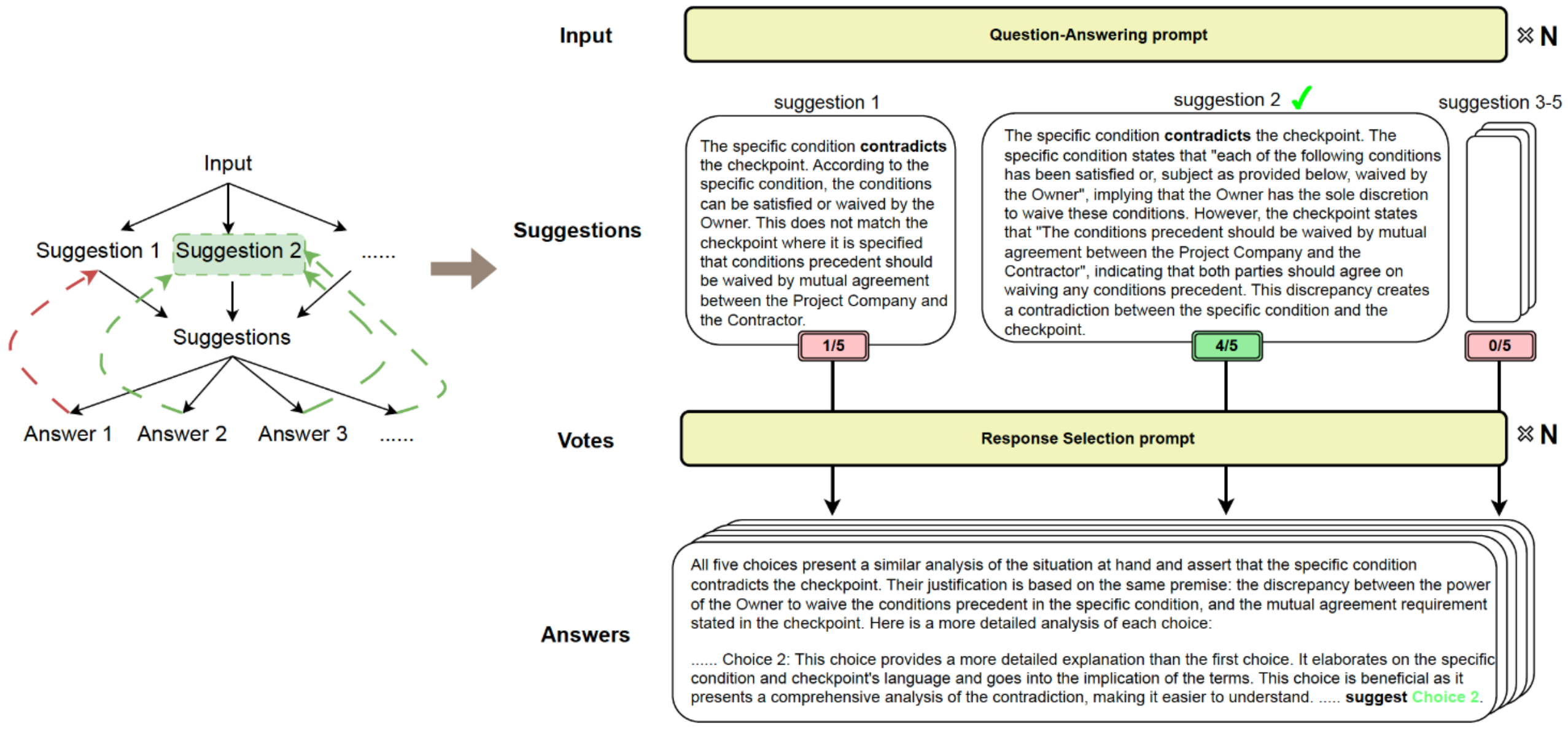

**Fig. 4**. An illustration of the two-stage prompting

Fig.5 shows the overall structure of the Question-Answering prompt. In tackling the challenge of LLMs struggling to determine the appropriate perspective or logical thinking for identifying relevant risks, we provide clause-review pairs as input-output exemplars. These exemplars serve as a guide for the LLMs' thinking process, employing the method of few-shot prompting instead of zero-shot.

**Instruction:** [Here are the checkpoint, specific conditions, and the relevant review…… the checkpoint.]

**Input-output exemplars:** [Input: {Context: <Checkpoint, Specific condition>,
Question: <Choose from the following options……>}],
[Output: {Explanation …… the answer is therefore}]

**Summary:**

**Input context:** [Checkpoint, Specific condition]

**Question:** [Choose from the following options……]

**Answer:** [Explanation …… the answer is therefore.]

**Fig. 5.** The structure of Question-Answering prompt.

The details of the Question-Answering prompt and Response Selection prompt are illustrated in Fig. 6. We have opted for multiple-choice questions as the format for our questions. If the answer is "contradict" or "not found," we consider the given particular condition as risky. Here, "contradict" implies that the specific condition is in contradiction with the checkpoint, while "not found" indicates potential risks to the integrity of the clauses. The contents within curly brackets, such as clause_1 or review_1, represent the information retrieved during the project clauses retrieval and clause-review pair retrieval steps. After providing exemplars, we requested the LLMs to concisely summarize the given information concisely and complete the subsequent tasks based on the summary, which helps LLMs to better understand the intention of the exemplars by summarizing what was provided or learned. Conversely, the Response Selection prompt is a standard prompt comprised of an Instruction and an Input context, where checkpoint and specific condition are identical to the Question-Answering prompt, and “choice i” denotes the result of the previous step.

| | **Prompt** |
|---|---|
| **Question-Answering** | Here are the checkpoint, specific condition, and the relevant review. Summarize what have you learned from the exemplars first concisely, no need to repeat the new question in your answer, and strictly follow the logic of the summary to answer the new question.<br>Checkpoint: ```{**checkpoint_A**}```<br>Specific condition: ```{**clause_1**}```<br>Choose from the following options: [A] contradict [B] entail [C] not found<br>Explanation: ```{**review_1**}```<br>The answer is therefore<br>Checkpoint: ```{**checkpoint_A**}```<br>Specific condition: ```{**clause_i**}```<br>Choose from the following options: [A] contradict [B] entail [C] not found<br>The answer is therefore<br>Explanation: ```{**review_i**}```<br>---<br>Summary: <Summarize here><br>---<br>Checkpoint: ```{**checkpoint_A**}```<br>Specific condition:```{**specific_conditions**}```<br>Choose from the following options: [A] contradict [B] entail [C] not found<br>Explanation:<br>The answer is therefore |
| **Response Selection (voting)** | Analyze choices below in detail, then conclude which is the most promising for the instruction and must giving one choice at the end.<br>Instruction: based on the specific condition, decide whether it contradicts with and entails with the checkpoint.<br>checkpoint:```{**checkpoint_A**}```<br>specific condition: ```{**specific_conditions**}```<br><br>choice 1: ```{**suggestion_1**}```<br><br>choice 2: ```{**suggestion_2**}```<br><br>choice 3: ```{**suggestion_3**}```<br><br>choice i: ```{**suggestion_i**}``` |

**Fig. 6.** The framework of Question-Answering and Response Selection prompt

According to (Yao et al., 2023), we can formalize the above intact processes for into formula (1) and (2). We use (1) to illustrate the procedure of acquiring suggestions using few-shot prompting with various specific conditions, where $P_\theta$ represent pre-trained language model with parameters $\theta$, in other words a LLM, $x$ as the input, and $y_i$ as the intermediate outputs through LLM with $i$ times (i = 1, 2, …, n). As mentioned in (Wang et al., 2022), more samples will result in a consistently better performance. We recommend the value of $i$ is at a minimum of 5. $prompt_{IO}(x)$ denote an input-output prompting which the input x involves instructions of the risk identification task, i.e., task description, input-output exemplars of Question and Answer pairs form as demonstrations for the LLM, and the input context, namely checkpoint and specific conditions.

$$y_i \sim P_\theta\big(y_i \mid prompt_{IO}(x)\big) \tag{1}$$

After getting $y_i$, we use (2) to acquire the final output $\hat{y}$, i.e. the final answer,

through selecting the most reasonable intermediate output from a set of $y_i$, where Y represent a set of $y_i$, and $prompt_{SEL}(x, Y)$ denote that a selection prompting which the input x only contains the task description and the input context.

$$\hat{y} \sim P_\theta\big(\hat{y} \mid prompt_{SEL}(x, Y)\big) \quad (2)$$

# 4. Case illustration

This section present examples that showcase the effectiveness of our approach in comparison to the standard prompt. Our approach involves using clause-review pairs as the expert knowledge and the context of the clauses requiring risk identification as the factual knowledge along side our prompts. The standard prompt, on the other hand, was developed according to instructions provided by OpenAI's official resources, only relying on the factual knowledge. All responses are generated using the GPT4.0 model[3].

## 4.1 Data preparation

We adopted EPC contracts collected from construction projects as our test data. One hundred clauses are extracted from the sample contracts and used for testing on the designated checkpoints, with 50 clauses from each. These selected clauses either contained information that need verification or had superfluous information not required for the checkpoints. The first and second contracts contain 22 and 7 clauses, respectively, relating to the designation of checkpoints. In addition, we prepared 20 clause-review pairs from other contracts, and the corresponding review results are provided by human experts.

The checkpoints used in this demonstration are: "The Financial Closing Date shall have occurred before the Commencement Date" and “The conditions precedent should only be waived by mutual agreement between the Project Company and the Contractor”. Further, the specific conditions used for this risk identification are composed of two parts: one consists of clauses related to the main elements of the checkpoint, namely the Commencement Date, Financial Closing Date, and Condition Precedents; the other part consists of a randomly selected assortment of clauses unrelated to the checkpoint. Furthermore, the clause-review pairs are put together by contract experts who identify relevant clauses from other commercial contracts based on the above checkpoints first, then provide the corresponding review results. All the clauses and reviews are saved in CSV format, following the data schema mentioned in section 3.1. For more details, please refer to this repository[4].

## 4.2 Data storage and retrieve

One way of leveraging the knowledge base to store and retrieve clauses data is to deploy it through a Docker in local environment. We utilized Milvus[5], an open-source and highly scalable vector database that supporting vector embeddings, to facilitate the storage and to index the massive vector embeddings. 48 specific conditions and 8 case clause-review pairs associated to the corresponding checkpoints were saved as the

[3] https://openai.com/gpt-4

[4] https://github.com/HarrisSK/Knowledge-augmented-LLMs-for-construction-contract-risk-identification/

[5] Milvus is accessible at: https://milvus.io/.

“specific_conditions” and “clause_review” respectively into three collections, and the data schema as well as the unit of text followed section 3.1. Since we use OpenAI's embedding model, we set the dimension of the embedding to its output value of 1536.

To retrieve related specific conditions and case clause-review pairs, Milvus provides an efficient indexing approach, HNSW (Hierarchical Navigable Small World Map), to rapidly access the target location, as well as supports the hybrid search. We set the index building and search parameters to its recommended values (i.e. Maximum degree of the node = 48, Search scope = 500). After having all the parameters set, we implement Euclidean distance metric to compare the checkpoint to each clause as well as to the retrieved specific conditions to each case clause (as illustrated in Fig. 7 and Fig. 8). We have extracted top 5 clauses out of “specific_condition_1” collection and top 3 clause-review pairs out of “clause_review” collection according to the Euclidean distance score.

**ALGORITHM 1: SPECIFIC CONDITIONS RETRIEVAL**

```
Input: P ← checkpoint,
       K ← the value to retrieve the top k relevant clauses,
       CB ← the clauses database
Initialize: similarityScores ← [], clauseSentence ← [], top_k_indices ← [],
top_k_clauses ← []
Set up API
vector_checkpoint = Langauge_Model.embed(P);
for vector_clause, i in CB
do
    similarityScores = calcualate_simularity(vector_checkpoint, vector_clause);
    similarityScores.append(similarityScores, i);
    similarityScores.sort(reverse ← True, x:x[0]);
    for source_tuple in similarityScores[K] do
        top_k_indicies = source_tuple[1];
    end
    for index in top_k_indices do
        top_k_clauses = CB.retrieve(index);
        clauseSentence.append(top_k_clauses);
    end
end
```

**Fig. 7.** Project clauses retrieval algorithm.

**ALGORITHM 2: CLAUSE-REVIEW PAIRS RETRIEVAL**

```
Input: P ← checkpoint,
       C ← clauseSentence,
       K ← the value to retrieve the top k relevant clauses,
       CCB ← the case clauses database
Initialize: relevant_clause_review_pairs ← [], top_k_clause_review_pairs ← [],
similarityScores ← [], top_k_indices ← []
Set up API
vector_clauseSentence = Language_Model.embed(clauseSentence);
for clause_review_pairs in CCB
do
    if P = CCB_checkpoint(i) then
        relevant_clause_review_pairs.append(clause_review_pairs);
    end
    for vector_clause, i in relevant_clause_review_pairs do
        similarityScores = calculate_simularity(vector_clauseSentence,
        vector_clause);
        similarityScores.append(similarityScores, i);
        similarityScores.sort(reverse ← True, x:x[0]);
        for source_tuple in similarityScores[K] do
            top_k_indices = source_tuple[1];
        end
        for index in top_k_indices do
            top_k_clause_review_pair = CCB.retreive(index);
        end
    end
end
```

**Fig. 8.** Clause-review pairs retrieval algorithm.

## 4.3 Identification result comparison

After conducting the necessary steps to determine the specific conditions required for identification, we proceeded to compare the performance of our prompt architecture with and without its use. When not using our architecture, we followed the official OpenAI prompt engineering instruction[6] and created standard prompts that included basic element such as instruction, context, output format, etc. The structure of the standard prompt is depicted in Fig. 9. All identifications were carried out in new conversations, without any historical conversations provided. This means the GPT-4 model did not have access to any information that may interfere with the task. In the case of using our prompt method, we initially posed the same query five times, utilizing three different clause-review pairs. Then, we apply the voting prompt, executing the same process five times based on the results of the initial queries. To obtain a range of suggestions, we recommend using a temperature value around 0.2-0.4 if temperature sampling is employed. The outcome with the highest number of votes was recognized as the risk identification result for the clause. For a fair comparison, we also used the standard prompt five times for the same query. The entire conversation is available in the repository.

[6] The official instructions for prompt engineering with ChatGPT: https://platform.openai.com/docs/guides/gpt-best-practices

| Prompt | |
|---|---|
| Construction contract risk identification | You are a construction contract review AI assistant.<br>You will be provided with checkpoint, specific condition delimited by ```.<br>Your task is to perform the following actions:<br>1 - check whether the specific condition contradicts with or entails with the checkpoint.<br>2 - if there are not enough information to determine whether the specific condition contradicts with or entails with the checkpoint, simply output not found.<br>3 - give explanation for your response.<br>Use the following format:<br>Condition situation: <contradict or entail or not found><br>Explanation: <Explanation><br><br>checkpoint: {checkpoint}<br>specific condition: {specific_conditoin} |

**Fig. 9.** Architecture of standard prompt for CCRI.

Table 2 and Table 3 presents the risk identification results of the two clauses for checkpoint 1 and checkpoint 2, respectively. The left-hand side is the response using standard prompt, while the right-hand side shows the response of our method. The identification results from both cases demonstrate that our method alignment with the expert reviews. For "Clause 1," the identification result should be that it is contradictory to the “Checkpoint 1”, as the Financial Close can be waived by the Employer, potentially resulting in the Financial Closing Date not occurring before the Commencement Date. For "Clause 2," the identification result should be that it entails with the “Checkpoint 2”, as the statement in “Clause 2” allows a party to unilaterally waive certain provisions, but these provisions are limited to those protecting its own rights and interest, without harming the other party. Thus, it does not violate the requirement for mutual consent to waive a condition precedent.

The result from identifying “Clause 1” and "Clause 2” failed to align with those of the human expert reviews when using standard prompt and not providing the clause-review pairs. In the case of “Clause 1”, the LLM only grasped the overall clause structure without a deep analysis of its content. The risk associated with this clause, which allows the Employer to waive the Financial Close, was not identified, leading to the incorrect conclusion that “Clause 1” entails with the “Checkpoint 1”. Similarly, for “Clause 2”, the LLM only assessed compliance with the “Checkpoint 2” based on the logic of “if the waiver is made without mutual consent.” Contradictions with the “Checkpoint 2” arise when the sentence contradicts this logic. The issue here is that the LLM fails to fully grasp the intended meaning of the clause.

**Checkpoint 1:** The Financial Closing Date shall have occurred before the Commencement Date.
**Clause 1:** "Commencement Date" means the date on which the conditions below have been satisfied or waived in accordance with the following sentence. The below condition (i) may be waived by the Employer under the condition that Contractor has received payment of the milestone defined as payment no. 2 in Schedule 4 (as down-payment, i.e. regardless of the conditions established for milestone no. 2), the below conditions (iii) and (vi) may be waived by the Employer and the below conditions (ii), (iv) and (v) may be waived by the Contractor:
(i) Financial Close;
(ii) the Contractor has received the down-payment from the Employer in the amount of 5% of the Contract Price as established as payment no,1 in Schedule 4;
(iii) not used; …

| | |
|---|---|
| **Standard:** Condition situation: Entail. Explanation: The specific condition does entail the checkpoint. The checkpoint states that the Financial Closing Date should occur before the Commencement Date. In the specific condition, it is detailed that the "Commencement Date" is the date on which certain conditions have been satisfied or waived, and one of those conditions is the Financial Close. This essentially means the same thing as the Financial Closing Date. Thus, it is clear that the Financial Closing Date (or the Financial Close) is intended to occur before the Commencement Date, which is in alignment with the checkpoint. | **Ours:** The specific condition contradicts the checkpoint. According to the provided specific conditions, the requirement for Financial Closing (stated as condition (i)) can be waived by the Employer under certain circumstances. Therefore, it is not definite that the Financial Closing Date will occur before the Commencement Date. The answer is therefore [A] contradicts. |

**Table 2.** An excerpt of risk identification result for checkpoint 1.

**Checkpoint 2:** The conditions precedent should only be waived by mutual agreement between the Project Company and the Contractor.
**Clause 2:** "Effective Date" means the date upon which the EPC Contractor receives a notice from the Owner requiring the EPC Contractor to proceed with the Works (the "Full Notice to Proceed"). The Full Notice to Proceed shall not be issued until each of the following conditions precedent has been satisfied, or waived by the Owner (in the case of articles 5.1(a), 5.1(b) and 5.1(e) of this Agreement), by the EPC Contractor (in the case of articles 5.1(c) and 5.1(f) of this Agreement) and by both Parties (in the case of articles 5.1(d) and 5.1(g) of this Agreement): (a) the Owner has received a certificate in respect of each of the following entities signed by a duly authorised officer of such entity:…

| | |
|---|---|
| **Standard:** Condition situation: Contradict. Explanation: The checkpoint condition specifies that the conditions precedent should be waived by mutual agreement between the Project Company and the Contractor. However, in the specific condition, the conditions precedent are to be waived by the Owner for articles 5.1(a), 5.1(b), and 5.1(e), by the EPC Contractor for articles 5.1(c) and 5.1(f), and by both parties for articles 5.1(d) and 5.1(g). This creates a contradiction as the parties involved in waiving the conditions precedent in the specific condition do not align with the party involvement in the checkpoint condition. | **Ours:** Explanation: The specific condition entails the checkpoint. The condition explicitly states that certain conditions can be waived by the Owner (representing the Project Company), others can be waived by the EPC Contractor, and a few can be waived by both parties. This distribution of authority aligns with the checkpoint's requirement that waivers should be made by mutual agreement, ensuring the interests of both parties. However, it also allows for individual waivers where the conditions solely concern the party providing the waiver, ensuring no compromise on their interests. The answer is therefore [B] entails. |

**Table 3.** An excerpt of risk identification result for checkpoint 2.

## 5. Discussion

The proposed method aims to develop an automatic pipeline for contract review and risk detection based on knowledge augmented LLM. We demonstrated the proficiency of our method with selected practical construction contracts. In order to adopt our method in contracts of a wider domain, we offer some insights and apprehensions that may arise during the implementation of the method in real-world scenarios.

### 5.1 "Logical thinking" of LLM

We observe from our experiment that accurate risk review conclusion comes with subtle human intervention. It is crucial to note that LLM has a higher tendency to follow the logical reasoning of human-provided demonstrations. Especially in contract review domain which requires higher level of alignment with experts, it is essential to carefully provide the demonstrations (clause-review pairs as in this paper).

Based on our experimental observations, there are two primary factors influence the efficacy of providing the clause-review pair to the LLM in emulating the thinking pattern of the contract experts, namely the quantity of clause-review pairs and the degree of similarity between the case clause and the risk clause. Regarding the former, when the provided case clause closely resembles the clause that requires scrutiny, both a single clause-review pair and three pairs can prompt the LLM to emulate the expert's thinking pattern in assessing risk clauses, implying that the LLM will review related clauses based on similar risk preferences. We also observed that even if discrepancies exist in the risk rationale for the same checkpoint among multiple similar clause-review pairs, the LLM's ability to mimic the experts' thinking pattern remains. As for the latter, when only one clause-review pair is provided and the case clause within that pair is not highly similar to the risk clause needing examination, the clause-review pair does not sufficiently stimulate the LLM to adopt the appropriate analytical mindset. Taking clause 1 from the case illustration as an example, when a less correlated clause-review pair is provided and the Euclidean distance between the case clause and clause 1 is 0.31, the LLM fails to accurately identify the inherent risks in clause 1. This might be due to the provided clause-review pair emphasizes the risk arising from the absence of the description stating "The Financial Closing Date shall have occurred before the Commencement Date", rather than from the provision allowing the employer to waive, which results in the risk.

Risks in construction contracts frequently appear in sentences with such subtle descriptive variations. Identifying such risks requires the ability to apply specific knowledge to certain circumstances. It is important to note that while LLMs can mimic logical reasoning, they do not truly “understand” or “think” about the topics the same way humans do. In the first example presented in section 4, the occurrence of the risk is related to risk preference, and the difference in responses depends on whether external knowledge of risk preference is provided. For a general language model without specific risk preferences, the first answer is reasonable. The risk of this clause lies in the preference level of "or" in "satisfied or waived." In the response utilizing our method, we provide a review pair with similar clauses. The clause-review pair, serving as knowledge enhancement, emphasizes the risk preference for "or" and highlights the risk caused by the uncertainty brought about by "or" to some extent.

Humans excel at learning from examples. Confronted with an unfamiliar issue, we often reason and derive solutions by comparing it with familiar situations. The results of our method suggest that LLMs demonstrate a similar ability, leading some to postulate that they exhibit a form of “logical thinking”. However, the perceived "logical thinking" of LLMs is merely a result of recognizing and generating patterns from a massive dataset. While they lack human-like cognition and reasoning abilities, they can emulate logical reasoning to a significant extent based on the data and the appropriate paradigm, namely the provided knowledge and the prompt (P. Liu et al., 2023). Just as data inherently sets the upper bound of a model's potential, the inclusion of informed professional analysis and well-defined risk preferences shape the judgment of a language model in risk identification. As a result, the construction of a comprehensive and precise knowledge base is fundamental to this endeavor. Furthermore, the continuous accumulation of risk-related clauses and expert analyses through practice contributes to sustainably performance improvement.

Traditionally, supervised learning-based models predict outputs based on inputs. In contrast, prompt-based learning models use language models to predict the probability of a text sequence. In this new paradigm, the initial input undergoes transformation through a template, resulting in a "prompt" with unfilled slots. The language model then probabilistically fills these slots to obtain the final output. Such an approach, with well-crafted prompts, enables the language model to flexibly perform few-shot or zero-shot learning, and even emulate "logical thinking." Therefore, for the language model to produce trustworthy and human-interpretable responses, prompts with sufficient context, clarity, and specificity are essential, particularly for highly relevant knowledge in downstream tasks.

### 5.2 Evaluation criterion for CCRI tasks

Although the risk review using our method can greatly improve the contract review efficiency, there doesn't exist a comprehensive evaluation metrics for evaluating the LLM generated content for contract review task. The establishment of such evaluation metrics faces a major challenge due to the diverse and complex evaluation standard in real-world scenarios.

Currently, the evaluation primarily relies on the feedback of human experts, using it as the ground truth. The expert assesses the response based on its logic, complexity, accuracy, and traceability. Specifically, they determine whether the response actually refers to the provided clauses, if its logic is sound, and if it accurately identifies the point of contradiction. To improve the efficiency of evaluations and mitigate the human subjective bias, one promising approach is to establish an iterative, automated expert evaluation system. Potential methods include using a knowledge graph for explicit knowledge inference (Zhang et al., 2020; Zheng et al., 2023). The identification results are compared with inference outcomes obtained from the knowledge graph, enabling the evaluation of answers generated by large language models from a logical inference perspective. On the other hand, it is also feasible to combine multiple language models, allowing them to serve as an evaluator that assesses answers for accuracy based on specific criteria. Answers that deviate from the standards, along with the corresponding adjustment strategies, are returned to the language models responsible for identifying risks, which then re-generate them. For instance, Yang's team (Y. Liu et al., 2023) used the LLMs along with a particular prompting paradigm to act as the evaluator for assessing natural language generation outputs. Their results show that the proposed method outperforms all previous methods. Furthermore, the establishment of relevant benchmark datasets is crucial for comprehensive performance evaluation of these tasks. Examples of benchmark datasets in other domains can be referenced in the works of (Chalkidis et al., 2022).

## 6. Conclusion

The paper presents a knowledge augmented language model methodology for CCRI, which integrates construction contract knowledge into LLMs. This approach incorporates domain-specific knowledge, contract clauses, and expert reviews to improve the efficiency and accuracy of contract-specific knowledge retrieval. The experiment results demonstrate the effectiveness of the method and provide valuable perspectives for discussion.

While the presented method performs well in the experiment, it has not yet been implemented in real-world scenarios that may involve a larger domain knowledge base. On one hand, a larger knowledge base could enhance the language model's comprehension by providing more prior knowledge. On the other hand, it may pose challenges in identifying the best few-shot learning materials. Therefore, it is recommended to collaborate with industrial practitioners to further investigate this direction.

## 7. Data availability

All data are available at https://github.com/HarrisSK/Knowledge-augmented-LLMs-for-construction-contract-risk-identification/.

## Acknowledgement

This research was funded by the National Natural Science Foundation of China (grant no. 71971196).